# Effective and Extensible Feature Extraction Method Using Genetic Algorithm-Based Frequency-Domain Feature Search for Epileptic EEG Multi-classification

Tingxi Wen[a], Zhongnan Zhang[a*]

[a] Software School, Xiamen University, Xiamen, Fujian, China 361005

Email: zhongnan_zhang@xmu.edu.cn

**Abstract**

In this paper, a genetic algorithm-based frequency-domain feature search (GAFDS) method is proposed for the electroencephalogram (EEG) analysis of epilepsy. In this method, frequency-domain features are first searched and then combined with nonlinear features. Subsequently, these features are selected and optimized to classify EEG signals. The extracted features are analyzed experimentally. The features extracted by GAFDS show remarkable independence, and they are superior to the nonlinear features in terms of the ratio of inter-class distance and intra-class distance. Moreover, the proposed feature search method can additionally search for features of instantaneous frequency in a signal after Hilbert transformation. The classification results achieved using these features are reasonable; thus, GAFDS exhibits good extensibility. Multiple classic classifiers (i.e., $k$-nearest neighbor, linear discriminant analysis, decision tree, AdaBoost, multilayer perceptron, and Naïve Bayes) achieve good results by using the features generated by GAFDS method and the optimized selection. Specifically, the accuracies for the two-classification and three-classification problems may reach up to 99% and 97%, respectively. Results of several cross-validation experiments illustrate that GAFDS is effective in feature extraction for EEG classification. Therefore, the proposed feature selection and optimization model can improve classification accuracy.

## 1. Introduction

Epilepsy, which leads to short-term brain dysfunction, is a chronic disease generated by a sudden abnormal discharge of brain neurons. In 2013, over 50 million patients were afflicted with epilepsy worldwide, with most patients originating from developing countries [1]. Approximately 9 million epileptic patients were recorded in China in 2011. Even worse, 600,000 new epileptic patients are recorded every year [2]. In China, epilepsy has become the second most common nerve disease, only second to headache. Therefore, accurate diagnosis and prediction of epilepsy are significant. Electroencephalogram (EEG) signal is often used to evaluate the neural activities of the brain. EEG signals, which are acquired by electrodes in the cerebral cortex, can reflect the states of brain neurons with time. The recorded EEG signals are complex, nonlinear, unstable, and random due to the complex interconnection between billions of neurons. Several scholars have focused on EEG signal analysis and process to aid in the diagnosis and treatment of epilepsy.

The first step in EEG signal analysis is to extract and select features. The major signal feature extraction methods are based on time-domain, frequency-domain, time–frequency domain, and nonlinear signal analyses [1]. Altunay et al. [3] presented a method for epileptic EEG detection based on time-domain features. Chen et al. [4] extracted features of EEG signals by using Gabor transform and empirical mode decomposition, which involves frequency-domain and time–frequency domain technologies. For nonlinear signal analysis, Zhang et al. [5] extracted six energy features and six sample entropy features of EEG signals. In feature extraction, researchers have often mixed multiple methods and obtained new features by various models. Zhang et al. [6] combined an autoregressive model and sample entropy to extract features, and results showed that the combination strategy could effectively improve the classification of EEG signals. Geng et al. [7] used correlation dimension and Hurst exponent to extract nonlinear features. Ren et al. [8] applied convolutional deep-belief networks to extract EEG features. By contrast, other researchers have extracted fixed features. Chen et al. [9] extracted dynamic features by recurrence quantification analysis. Tu and Sun [10] proposed a semi-supervised feature extractor called semi-supervised extreme energy ratio (SEER). Improving on this work, they further proposed two feature extraction methods: 1) semi-supervised temporally smooth extreme energy ratio (EER) and 2) semi-supervised importance weighted EER [11]. These two methods presented better classification capability than SEER. Rafiuddin et al. [12] used wavelet-based feature extraction, statistical features, inter-quartile range, and median absolute deviation to form the feature vector. Wavelet packet decomposition was also used to extract EEG features [13].

After feature extraction, the selected features should be classified to recognize different EEG signals. Numerous classifiers are used for EEG classification, and they can be divided into five categories, namely, linear classifiers, neural networks, nonlinear Bayesian classifiers, nearest neighbor classifiers, and combinations of classifiers [14]. Li et al. [15] used a support vector machine (SVM) for multiple kernel learning to classify EEG signals. Murugavel et al. [16] also used an SVM but modified the machine into an adaptive multi-class SVM. Zou et al. [17] classified EEG signals by Fisher linear discriminant analysis (LDA). Djemili et al. [18] fed the feature vector to a multilayer perceptron (MLP) neural network classifier. The classification capacity of a single classification method is limited; thus, an increasing number of researchers have attempted to combine two or more methods to improve classification accuracy. For example, Subasi et al. [19] adopted an artificial neural network (ANN) and logistic regression to classify EEG signals. Wang et al. [20] combined cross-validation (CV) with $k$-nearest neighbor ($k$-NN) to construct a hierarchical knowledge base to detect epilepsy. Murugavel et al. [21] also proposed a novel hierarchical multi-class SVM, with extreme learning machine as kernel, to classify EEG signals. To classify multi-subject EEG signals, Choi [22] used multi-task learning, which treated subjects as tasks to capture inter-subject relatedness in the Bayesian treatment of probabilistic common spatial patterns.

Researchers have also studied the application of machine learning and optimization algorithms to improve the accuracy of epilepsy detection. Amin et al. [23] compared the classification accuracy rate of SVM, MLP, $k$-NN, and Naïve Bayes (NB) classifiers for epilepsy detection. Nunes et al. [24] used the optimum path forest classifier for seizure identification. Moreover, artificial bee colony [25] and particle swarm optimization [26] algorithms were also used to optimize neural networks for EEG data classification. However, the study of the application of machine learning and optimization algorithm to epilepsy detection is currently insufficient.

In this paper, a genetic algorithm-based frequency-domain feature search (GAFDS) method is proposed. This method searches for better classification features in the frequency spectrum rather than using the maximum, minimum, and mean values of the frequency spectrum as features. This method can be easily extended to the feature extraction of other spectra. As for the multi-classification problem, a high classification accuracy can be achieved by using the features acquired by the GAFDS method. In addition, the accuracy can be further improved by combining other nonlinear features. In addition, an optimization algorithm is used to optimize feature selection.

The remainder of this paper is organized as follows: Section 2 describes the feature extraction methods, including the feature selection and classification models. Section 3 presents the experimental dataset, process, and results. Section 4 discusses the experiments. Finally, Section 5 concludes the paper.

## 2. Methodology

The EEG dataset used in this study is obtained from a public dataset. As shown in Fig. 1, numerous features of the EEG signals were first extracted by various feature extraction methods. Subsequently, the best feature combination subset is selected by a feature selection method. Finally, the feature combination subset is used by a classifier for the classification of the EEG signals.

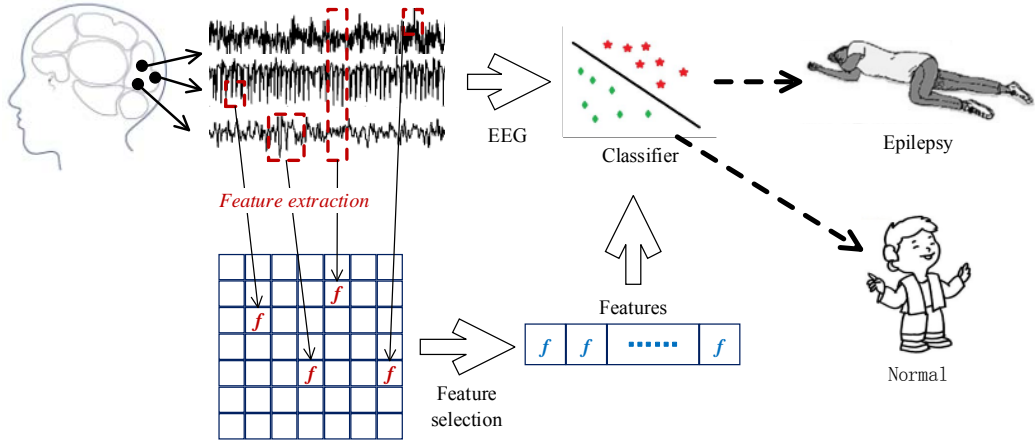

Fig. 1 Overall process of EEG signal classification

### 2.1 Feature Extraction
#### 2.1.1 Genetic algorithm-based frequency-domain feature search method (GAFDS)

Genetic algorithm (GA) is a random search method that simulates the biological laws of evolution. In this study, GA adopts the probability optimization method and exhibits global optimization capability. The standard parallel GA [27] is used to search for features in frequency domain.

$$GA = \{C, E, P_0, M, \Phi, \Gamma, \Psi, T\}, \tag{1}$$

where $C$ is the chromosome coding in GA, $E$ is the individual fitness function, $P_0$ is the initial population, $M$ is the size of the initial population, $\Phi$ is the selection operator, $\Gamma$ is the crossover operator, $\Psi$ is the mutation operator, and $T$ is the given termination condition.

In signal processing, the frequency domain is a coordinate system that describes the frequency features of signals. Often used to analyze signal features, a frequency spectrogram reflects the relationship between the frequency and amplitude of the signals. The GAFDS method adopts GA to search for a set of classification-suitable features in the frequency spectrum.

Fig. 2 shows the frequency spectrogram of five classes of signals (i.e., A, B, C, D, and E) after fast Fourier transformation (FFT). The *x*-axis is the frequency, whereas the *y*-axis is the amplitude. A significant variation occurs in each class when the frequency is at a certain value, such as the amplitudes in the red rectangle frames. However, the amplitudes in the green rectangle frame are difficult to distinguish. The proposed feature extraction method searches for several superior frequency value spaces in the frequency spectrogram. Subsequently, the mean values of the amplitudes in the spaces are used as the features. Next, the GA with global search capability is employed to search for frequency value spaces.

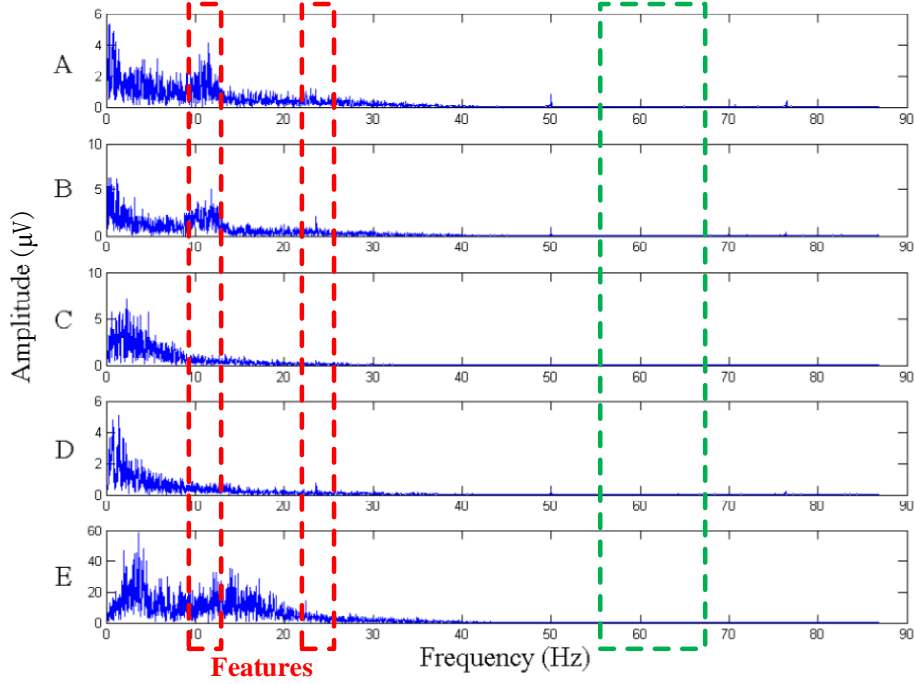

Fig. 2 Feature extraction based on the frequency domain

A time series $X\{x_1, x_2, \ldots, x_n\}$ with a length of *n* is formed after a signal is sampled. Then, a series $Y\{y_1, y_2, \ldots, y_m\}$ with a length of *m* is obtained by applying FFT to *X*. For $i, j \in [1, \ldots, m]$ and $i < j$,

$$f_{ij} = \frac{1}{j-i+1}\sum_{k=1}^{j} y_k. \qquad (2)$$

The $f_{ij}$ in Eq. (2) is the feature in the frequency interval $[i, j]$.

The main process of using GA in frequency intervals involves obtaining several frequency features with significant distinguishing capability. Details of this process follow.

1. Individual encoding

On the assumption that the total number of features to be searched for is $\alpha$, the length of the individual coding array *C* is $2\alpha$. The value of each element in *C* is between zero and the largest frequency. The two values $C_{2i}$ and $C_{2i+1}$ $(0 \le i < \alpha)$ from *C* are taken as the frequency range to calculate the features. As shown in Fig. 3, $C_3$ and $C_4$ can be used to calculate the feature $f_{C_3, C_4}$.

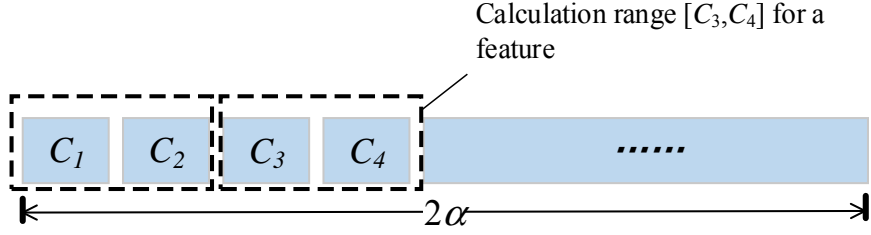

Fig. 3 Structure of the individual encoding

However, the constraint $i < j$ should be applied to calculate feature $f_{i,j}$. When $i \geq j$, the feature makes no sense. Therefore, a negative slack variable $\beta$ (when $i \geq j$, $\beta = 0$) is adopted to implement the constraint.

2. Fitness function

Traversing $C$ to calculate features yields $\alpha$ features $\{f_g^1, f_g^2, \ldots, f_g^\alpha\}$ and $\alpha$ slack variables $\{\beta_1, \beta_2, \ldots, \beta_\alpha\}$. For the optimization of features, the samples should ideally present larger inter-class distances and smaller intra-class distances in the feature space. LDA is employed to evaluate these features. The calculation involves a large number of iterations, and LDA has high calculating speed. The calculation of fitness value is

$$fitness(C) = \text{LDA}(f_g^1, f_g^2, \ldots, f_g^\alpha) - \sum_{i=1}^{\alpha} \beta_i. \qquad (3)$$

3. Operators

In GAFDS, $\Gamma$ is a multipoint crossover operator, $\Psi$ is a Gaussian mutation operator, and $\Phi$ is a roulette wheel selection operator.

**2.1.2 Nonlinear features**

An EEG signal is random and unstable; thus, using FFT alone cannot effectively distinguish EEG signals. Therefore, other nonlinear methods are used in this study to extract features, namely, sample entropy, Hurst exponent, Lyapunov exponent and multi-fractal detrended fluctuation analysis (MFDFA).

Proposed by Richman [28] in 2000, sample entropy improves on Pincus' approximate entropy [29], which is a measure of regularity to quantify the levels of complexity within a time series. Sample entropy is often used to extract features of EEG signals [5, 6]. The feature based on sample entropy is defined by

$$f_{se} = SampleEntopy(X, sn, sm, sr). \qquad (4)$$

Sample entropy requires three parameters: signal length, $sn$; embedding dimension, $sm$; and similar tolerance, $sr$. The value of $sr$ is the standard deviation of $X$ multiplied by parameter $\chi$.

The Hurst exponent was first proposed by England hydrologist H. E. Hurst [30]. It is often used in the chaos–fractal analysis of a time series and as the index for judging whether the time series data is random walk or biased random walk. In a previous study [7], the Hurst exponent is adopted as the main feature of EEG classification and defined by

$$f_{hurst} = HurstExponent(X). \qquad (5)$$

The Lyapunov exponent is used for computing how fast nearby trajectories in the dynamic system diverge. This exponent is one of the features used to recognize chaotic motions [31]. In this study, the largest Lyapunov exponent (LLE) is used as a feature of EEG, and it is given by

$$f_{lle} = LLE(Y). \qquad (6)$$

In physiology, fractal structures exist in physiological signals. Multi-fractals can reveal the complexity and inhomogeneity of fractals. MFDFA is the algorithm used for analyzing the multi-fractal spectrum of a biomedical time series [32].

Fig. 4 presents the MFDFA-based multi-fractal spectra of the five classes of signals (i.e., A, B, C, D, and E). When the $q$-order moments of the wave function are −8, −6, −4, −2, 0, 2, 4, 6, and 8, several multi-fractal spectra are formed. Three points from the multi-fractal spectrum, namely, $p_1$, $p_2$, and $p_3$, are selected, where $p_1$ is the point with the minimum $h_q$, $p_2$ with the maximum $h_q$, and $p_3$ with the maximum $D_q$. Each point has two coordinate values; thus, six features are obtained. The six features include $f_{mf}^1, f_{mf}^2$ (i.e., the $h_q$ and $D_q$ values of $p_1$); $f_{mf}^3, f_{mf}^4$ (i.e., the $h_q$ and $D_q$ values of $p_2$); and $f_{mf}^5, f_{mf}^6$ (i.e., the $h_q$ and $D_q$ values of $p_3$). The maximum value of $D_q$ is always equal to 1; thus, $f_{mf}^6$ can be removed. Therefore, the feature set $\{f_{mf}^1, f_{mf}^2, \ldots, f_{mf}^5\}$ can be obtained using MFDFA. At the same time, the detrended fluctuation analysis value of the signal is also a feature:

$$f_{dfa} = DFA(Y). \tag{7}$$

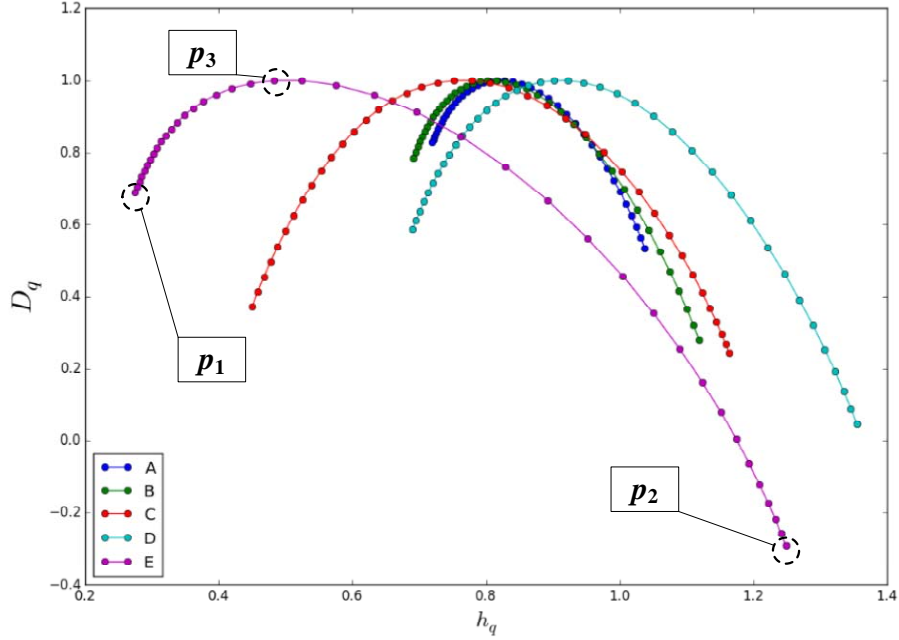

Fig. 4 MFDFA-based multi-fractal spectra of the five classes of signals

### 2.1.3 Feature selection and optimization

Feature extraction is useful in data visualization and comprehension. In addition, it reduces the requirement for data calculation and storage as well as the time for training and application. Numerous signal feature extraction algorithms are used in practice. Researchers often combine several feature extraction algorithms to analyze data. However, the use of multiple algorithms usually results in feature dimension expansion and feature redundancy. Feature selection reduces feature space dimension, thus facilitating data training and application.

The selection of an optimal feature subset is an NP problem; therefore, GA is used to search for the optimal feature subset. The algorithm codes individuals in the population in a binary array whose length is the number of features. In the array, 1 means the feature is selected, whereas 0 indicates otherwise. The object function of the algorithm is

$$min(\text{FPR} - (1 - \text{TPR})). \qquad (8)$$

where FPR is the fall-out or false positive rate and TPR is the sensitivity or true positive rate.

**2.2 Classification Model**

After feature extraction, multiple models, including k-NN, LDA, decision tree (DT), AdaBoost (AB), MLP, and NB, are used to classify EEG signals.

Cover and Hart [33] first proposed *k*-NN. The main idea of *k*-NN is as follows: if most of the *k* samples most similar (nearest in feature space) to a sample belong to a class, then the sample also belongs to that class.

LDA was introduced into pattern recognition and artificial intelligence by Belhumeur [34]. The basic functional concept of LDA is projecting high-dimensional pattern onto the optimal discriminant vector space to extract classification information and reduce feature space dimension. After projection, the pattern samples exhibit the maximum inter-class distances and the minimum intra-class distances in the new subspace, that is, the pattern presents the best separability in the space.

DT [35] implements a group of classification rules represented by a tree structure to minimize the loss function on the basis of the known occurrence probability of each situation. This model is a graphical method that intuitively uses probability analysis. The decision nodes resemble branches of a tree, and thus, it is called decision tree.

AB [36] is an iterative algorithm that trains different classifiers (weak classifiers) with the same training set and combines these weak classifiers with different weights to construct a stronger classifier (strong classifier).

MLP [37] is a feedforward ANN consisting of multiple layers of nodes in a directed graph, with each layer fully connected to the next. Each node is a processing element with an activation function. MLP uses backpropagation for training the network to distinguish data.

NB [38] is the classification method based on Bayes' theorem and feature conditional independence assumption. NB originates from classical mathematics and offers solid mathematical basis and stable classification accuracy. In addition, this model only requires few parameters, and it is not sensitive to missing data. In theory, the NB model presents less error than other classification methods do.

**3. Experiments and Results**

**3.1 Dataset**

The dataset is obtained from the work of Andrzejak et al. [39]. It includes five classes of data (i.e., A, B, C, D, and E). Each class has 100 single-channel EEG samples each with a length of 23.6 s. The sampling frequency is 173.61 Hz; thus, each sample is a time series with 4097 numbers. Sets A and B are collected when healthy volunteers open and close their eyes. Sets C, D, and E are from epileptics. The samples in set D are recorded from the epileptogenic zone, and those in set C are obtained from the hippocampal formation of the opposite hemisphere of the brain. Sets C and D contain only the activities measured during seizure-free intervals, whereas set E only contains seizure activity. The relation among these five classes of EEGs is shown in Fig. 5. Numerous previous studies focused on A, E classification; {C, D}, E classification; A, D, E classification; and A, B, C, D, E classification. This paper examines A, E classification; {C, D}, E classification; and A, D, E classification. Fig. 6 illustrates the sample data of five classes.

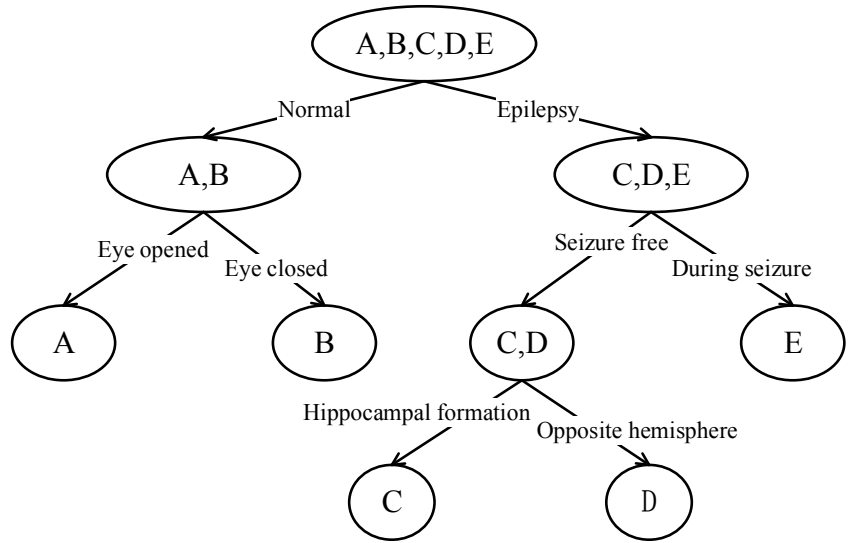

Fig. 5 Relation graph for the five classes of EEG signals

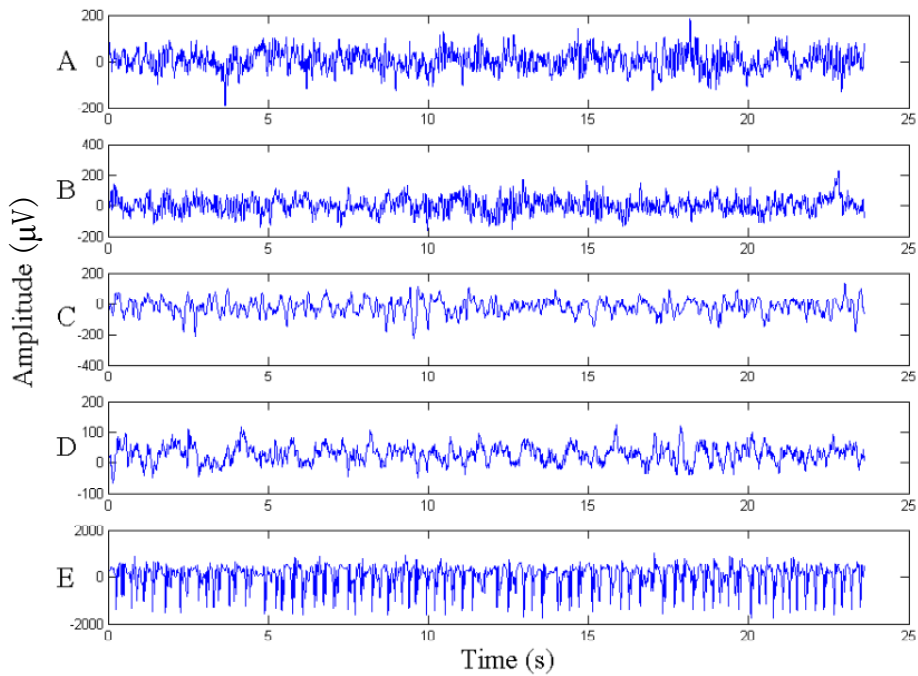

Fig. 6 Sample data of the five classes (i.e., A, B, C, D, and E)

**3.2 Results**

**3.3.1 Feature extraction and selection**

The feature selection results influence the classification results. The nonlinear features of signals do not change in different classification problems. However, the optimization variables of the object function of GAFDS vary across different classification problems, and the obtained features are different. In this study, the object function of GAFDS distinguishes the five classes. When $\alpha = 4$, four features (i.e., $f_1, f_2, f_3$, and $f_4$) are extracted by GAFDS, and five features (i.e., $f_5, f_6, f_7, f_8$, and $f_9$) are obtained by MFDFA. In addition, sample entropy $f_{10}$, Hurst exponent $f_{11}$, largest Lyapunov exponent $f_{12}$, and DFA feature $f_{13}$ are obtained.

Fig. 7 shows the distributions of the five classes of samples in feature space $f_1$. Most feature values of class A are below 3, whereas most features values of class E exceed 3. Therefore, classes A and E can be distinguished by feature $f_1$. Approximately 50% of the samples of classes B and C can be classified by $f_1$, but the other samples are mixed. Most class C and class D samples are mixed and difficult to distinguish. Overall, classes C and D present outliers, and class E is discrete. Fig. 8 shows the distributions of each class in different feature spaces. The median increases from class A to class E for $f_4$ but decreases for $f_6$. Feature $f_{13}$ can effectively distinguish {B, C} and {B, D}. This analysis shows that a single feature usually distinguishes two classes at most. Therefore, further analysis of the feature combination is necessary.

As shown in Fig. 9, features $f_1, f_4, f_6, f_7, f_9, f_{10}, f_{11}$, and $f_{13}$ are extracted to construct different two-dimensional feature spaces. A point in the space represents a sample of a class. In space $\{f_1, f_{11}\}$, all samples of the five classes are concentrated on the left side and difficult to separate. In space $\{f_4, f_9\}$, the outlines of classes A, B, E, and {C, D} are clear, but classes C and D are mixed. Every class is discrete in space $\{f_6, f_7\}$ and $\{f_{10}, f_{13}\}$; however, in space $\{f_{10}, f_{13}\}$, each class occupies a certain distribution area, and each class crosses only on the edges.

In one-dimensional or two-dimensional feature spaces, features can be directly observed. However, as the dimension increases, feature evaluation based on distances is the most direct method, regardless of the classifier. In this study, the features $\{f_1, f_2, f_3, f_4\}$ extracted by GAFDS are evaluated by comparing the ratios of inter-class distance and intra-class distance of all the classes with those of the nonlinear features $\{f_{10}, f_{11}, f_{12}, f_{13}\}$. The inter-class distance between two classes is their distance in a feature space. With $n$ samples in class A, $n$ feature vectors $\{v_1^A, \dots, v_i^A, \dots, v_j^A, \dots v_n^A\}$ are generated after feature extraction. Vector $v_i^A$ represents sample $i \in A$; thus, the intra-class distance of class A is calculated by

$$dist1 = \frac{1}{n \cdot n} \sum_{i=1}^{n} \sum_{j=1}^{n} (v_i^A - v_j^A)^2. \tag{9}$$

The distance between sample $i$ and itself is 0. The inter-class distance between $n$ samples in class A and $m$ samples in class B is calculated by

$$dist2 = \frac{1}{n \cdot m} \sum_{i=1}^{n} \sum_{j=1}^{m} (v_i^A - v_j^B)^2. \tag{10}$$

Therefore, the ratio of the inter-class distance and intra-class distance of class A and class B is

$$r_{AB} = dist2/dist1. \tag{11}$$

The ratio of the inter-class distance and intra-class distance of a class and itself is 1, that is, $r_{AA} = r_{BB} = r_{CC} = r_{DD} = r_{EE} = 1$.

Table 1 shows the ratios of inter-class distance and intra-class distance of the five classes in feature spaces $\{f_1, f_2, f_3, f_4\}$ and $\{f_{10}, f_{11}, f_{12}, f_{13}\}$. The features retrieved by GAFDS are comparable to the nonlinear features.

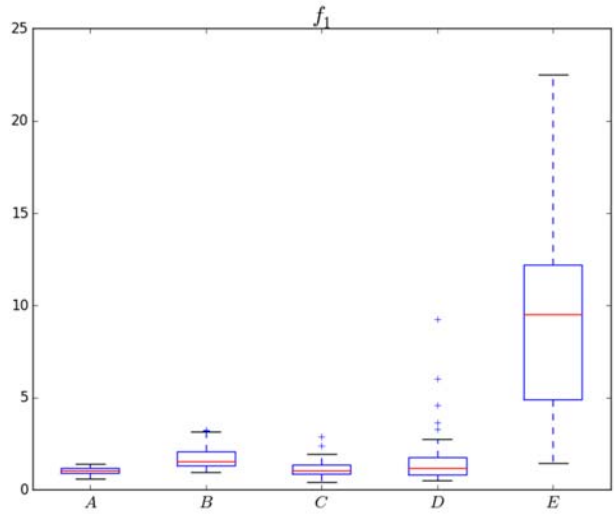

Fig. 7 Distributions in $f_1$ of the five classes of samples

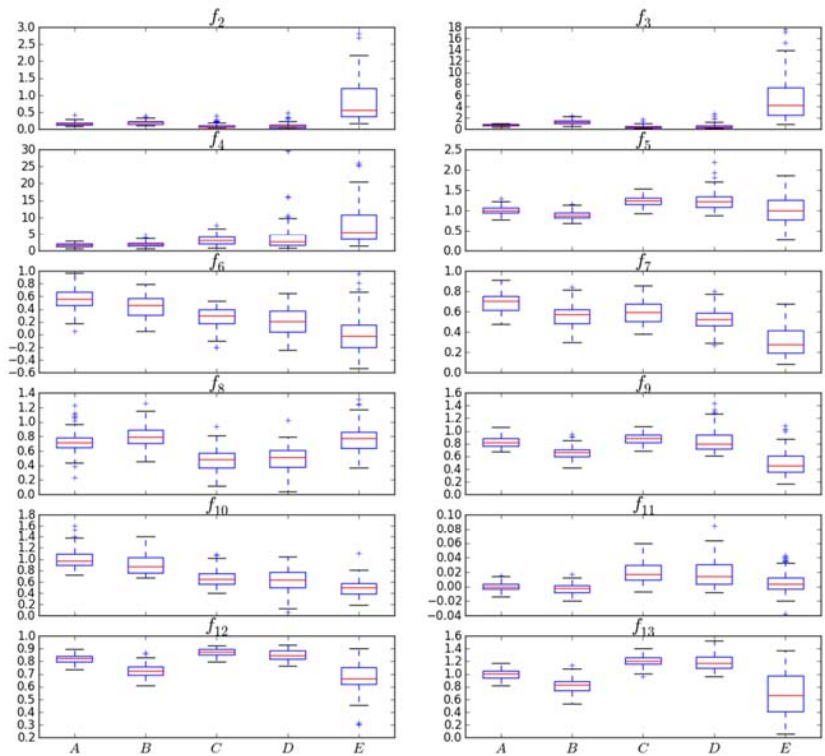

Fig. 8 Distributions in $f_1,…,f_{13}$ of the five classes of samples

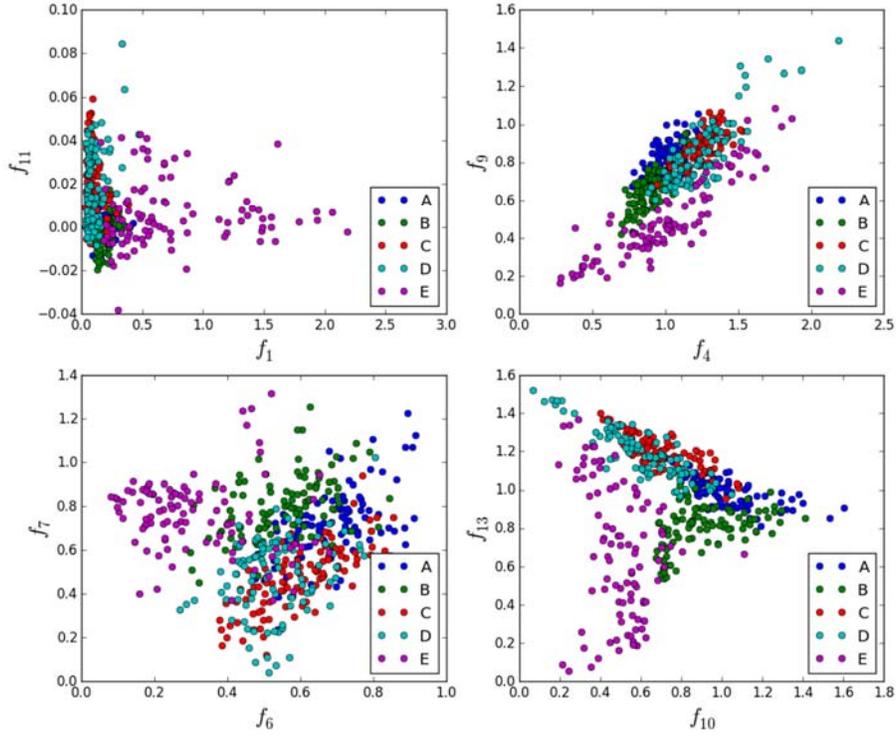

Fig. 9 Distributions of the five classes in different two-dimensional feature spaces

Table 1 Ratios of inter-class and intra-class distances of the five classes in feature spaces $\{f_1, f_2, f_3, f_4\}$ and $\{f_{10}, f_{11}, f_{12}, f_{13}\}$

|   | $f_1, f_2, f_3, f_4$ | | | | | $f_{10}, f_{11}, f_{12}, f_{13}$ | | | | |
|---|---|---|---|---|---|---|---|---|---|---|
|   | A | B | C | D | E | A | B | C | D | E |
| A | 1 | 0.745 | 0.826 | **2.996** | 0.743 | 1 | 0.914 | 1.952 | 2.355 | 1.642 |
| B | **1.793** | 1 | 1.318 | **7.067** | 1.110 | 1.238 | 1 | 2.616 | 3.296 | 2.061 |
| C | **2.434** | 1.613 | 1 | **8.405** | 1.009 | 1.980 | 1.958 | 1 | 2.125 | 1.233 |
| D | 1.176 | 1.153 | 1.120 | 1 | 1.198 | 1.368 | 1.413 | 1.217 | 1 | 1.521 |
| E | **4.250** | **2.639** | **1.959** | **17.460** | 1 | 2.166 | 2.006 | 1.603 | 3.452 | 1 |

The five classes of samples in $\{f_1, f_2, f_3, f_4\}$ and $\{f_{10}, f_{11}, f_{12}, f_{13}\}$ feature spaces are simultaneously classified by numerous classic classifiers.

Table 2 Classification accuracies of common classifiers for classes A and E in feature spaces $\{f_1, f_2, f_3, f_4\}$ and $\{f_{10}, f_{11}, f_{12}, f_{13}\}$ (using $k$-fold cross-validation, $k = 5$)

| **A, E** | **$k$-NN** | **LDA** | **DT** | **MPL** | **AB** | **NB** |
|---|---|---|---|---|---|---|
| $\{f_1, f_2, f_3, f_4\}$ | 0.995 | 0.89 | 0.995 | 0.995 | 0.995 | 0.995 |
| $\{f_{10}, f_{11}, f_{12}, f_{13}\}$ | 0.995 | 0.995 | 0.975 | 0.985 | 0.97 | 0.98 |

A, E classification involves classifying EEG signals produced by healthy people and epileptics.

As shown in Table 2, numerous classifiers can achieve high accuracies by using the listed features.

The classification of {C, D} and E means classifying EEG signals produced during seizure-free intervals and during seizure. GA is used to select the features for this classification. The results are shown in Table 3. A, D, E classification is the classification of EEG signals acquired from healthy people, seizure-free epileptics, and epileptics during seizure. The results are shown in Table 4.

Table 3 Classification accuracies of common classifiers for {C, D} and E classes in feature spaces $\{f_1, f_2, f_4, f_5, f_7, f_8, f_9, f_{10}, f_{12}\}$ (using $k$-fold cross-validation, $k = 5$)

| {C, D}, E | $k$-NN | LDA | DT | MPL | AB | NB |
|---|---|---|---|---|---|---|
| $\{f_1, f_2, f_4, f_5, f_7, f_8, f_9, f_{10}, f_{12}\}$ | 0.973 | 0.983 | 0.957 | 0.98 | 0.987 | 0.977 |

Table 4 Classification accuracies of the common classifiers for A, D, E in feature spaces $\{f_1, f_2, f_4, f_5, f_6, f_9, f_{10}, f_{11}, f_{12}, f_{13}\}$ (using $k$-fold cross-validation, $k = 5$)

| A, D, E | $k$-NN | LDA | DT | MPL | AB | NB |
|---|---|---|---|---|---|---|
| $\{f_1, f_2, f_4, f_5, f_6, f_9, f_{10}, f_{11}, f_{12}, f_{13}\}$ | 0.967 | 0.917 | 0.967 | 0.933 | 0.967 | 0.917 |

**3.3.2 Classification results**

In the previous section, the features extracted by GAFDS are optimized for the classification problem of five classes. However, for real binary or three-classification problems, the optimization object should be set according to the requirement, that is, the object function in Eq. (3) should be adjusted.

Table 2 shows good results for A, E classification. For the {C, D} and E classification problem, Table 5 lists the results based on the features extracted by GAFDS, whereas Table 6 illustrates the results based on the optimized features by GA selection from the GAFDS-obtained features and other features.

Table 5 Accuracies for {C, D}, E classification based on features

| Cross-validation | $k$-NN | LDA | DT | MPL | AB | NB |
|---|---|---|---|---|---|---|
| **2-fold** | 0.963 | 0.910 | 0.947 | 0.943 | 0.963 | 0.947 |
| **5-fold** | 0.967 | 0.907 | 0.963 | 0.953 | 0.973 | 0.950 |
| **10fold** | 0.950 | 0.923 | 0.973 | 0.96 | 0.970 | 0.950 |

Table 6 Accuracies for {C, D}, E classification based on the optimized features by GA selection

| Cross-validation | $k$-NN | LDA | DT | MPL | AB | NB |
|---|---|---|---|---|---|---|
| **2-fold** | 0.977 | 0.983 | 0.960 | 0.963 | 0.973 | 0.95 |
| **5-fold** | 0.983 | 0.983 | 0.973 | 0.97 | 0.98 | 0.957 |
| **10-fold** | 0.98 | 0.983 | 0.96 | 0.976 | **0.99** | 0.967 |

For a multi-classification problem, Tables 7 and 8 show that GAFDS and the feature selection method can obtain good results for A, D, E classification.

Table 7 Accuracies for A, D, E classification based on features extracted by GAFDS

| Cross-validation | $k$-NN | LDA | DT | MPL | AB | NB |
|---|---|---|---|---|---|---|
| **2-fold** | 0.93 | 0.857 | 0.917 | 0.9 | 0.693 | 0.81 |
| **5-fold** | 0.973 | 0.887 | 0.897 | 0.9 | 0.67 | 0.823 |
| **10-fold** | 0.937 | 0.886 | 0.92 | 0.913 | 0.71 | 0.817 |

Table 8 Accuracies for A, D, E classification based on the optimized features by GA selection

| Cross-validation | k-NN | LDA | DT | MPL | AB | NB |
|---|---|---|---|---|---|---|
| **2-fold** | 0.93 | 0.94 | 0.913 | 0.916 | 0.68 | 0.897 |
| **5-fold** | 0.96 | 0.953 | 0.927 | 0.94 | 0.587 | 0.907 |
| **10-fold** | 0.943 | 0.953 | 0.92 | 0.933 | 0.71 | 0.893 |

## 4. Discussion

### 4.1 GAFDS Method

EEG signals are nonlinear, time varying, and unbalanced. FFT is a global linear method. However, a frequency spectrum does not reflect the frequency changes in the time domain; thus, FFT has certain limitations when applied to non-stationary signal analysis. The features extracted by GAFDS present poor cohesiveness compared with other features. For example, class E presents a large distribution in $f_6$, whereas class D has numerous outliers. Thus, features $\{f_1, f_2, f_3, f_4\}$ in Figs. 7 and 8 are standardized to obtain new features $f_1^*, f_2^*, f_3^*, f_4^*$ within the range [0, 1]. After feature standardization, the accuracy of the AB classifier presents minimal reduction, whereas those of other classifiers remain unchanged, as shown in Table 9. These results indicate that the features extracted by GAFDS have better independence.

Table 9 Classification accuracies of common classifiers for A and E classes in feature spaces $\{f_1^*, f_2^*, f_3^*, f_4^*\}$ (using $k$-fold cross validation, $k = 5$)

| A, E | k-NN | LDA | DT | AB | MPL | NB |
|---|---|---|---|---|---|---|
| $\{f_1^*, f_2^*, f_3^*, f_4^*\}$ | 0.995 | 0.89 | 0.99 | 0.925 | 0.99 | 0.995 |

Table 1 presents a comparison based on Eq. (11) between the features extracted by GAFDS with nonlinear features. As shown in the table, $r_{AB}$ is good even though the features extracted by GAFDS present poor cohesiveness. Therefore, the features extracted by GAFDS are superior to the nonlinear features in A, E classification. Furthermore, GAFDS has great extensibility. GAFDS selects features by searching frequency spectrum; However, it can also search for new features in a Hilbert spectrum and several other signal spectra.

### 4.2 Analysis and Comparison of Classification Results

As shown in Tables 2–4, the classifiers have different accuracies in different feature spaces. This study uses less features and smaller searching space. The classification results show that the GA-based feature selection can obtain superior feature combination.

For the A, E classification problem, the features extracted by GAFDS can effectively facilitate classification. For the {C, D} and E classification, Tables 5 and 6 show that, after combining new features with the features extracted by GAFDS and feature selection, the classification accuracy increases. However, when the complexity of the problem increases, such as the A, D, E classification (Tables 7 and 8), the classification accuracies of the classifiers using the features extracted by GA selection are not significantly higher than those of the classifiers using the features only generated by GAFDS. Furthermore, the AB classifier performs well in the two-classification problem but poorly in the multi-classification problem because the parameters of the classifiers are not optimized. Using wavelet transform-based statistical features, largest Lyapunov exponent, and approximate entropy features, Murugavel et al. [21] developed an ANN and hierarchical multi-class SVM with a new kernel classifier to improve A, D, E classification accuracy to 96%. Sharma et al. [46] used the features based on two-dimensional and three-dimensional phase space representation of intrinsic mode functions, and SVM classifier to classify {C, D}, E and achieved an accuracy of 98.67%. The current work exhibits better classification results with several classifiers based on the GAFDS-selected features and nonlinear features.

## 5. Conclusion

EEG provides important information for epilepsy detection. Feature extraction, selection, and optimization methods exert significant influence in EEG classification. A GA-based frequency feature search method is proposed for EEG classification. The method features global searching capability to search for classification-aiding features in EEG frequency spectra and combine them with nonlinear features. Finally, GA is used to select effective features from the feature combination to classify EEG signals.

In the experiments, the standardization and normalization of the features extracted by GAFDS do not affect the accuracy of classification results, indicating that the features extracted by GAFDS have good independence. Compared with nonlinear features, GAFDS-based features allow for high classification accuracy. Furthermore, GAFDS can effectively extract features of instantaneous frequency in the signal after Hilbert transformation, suggesting the good extensibility of GAFDS.

For the A, E and {C, D}, E two-classification problems and the A, D, E three-classification problem, the GAFDS-based features and optimized features are used by several classifiers (i.e., $k$-NN, LDA, DT, AB, MLP, and NB). The classification accuracies achieved are better than those by previous classification models.